%% file: main.tex
\documentclass[10pt,twocolumn,letterpaper]{article}

\usepackage[pagenumbers]{cvpr} 
\usepackage{bm}
\input{preamble}

\definecolor{cvprblue}{rgb}{0.21,0.49,0.74}
\usepackage[pagebackref,breaklinks,colorlinks,citecolor=cvprblue]{hyperref}

\def\paperID{*****} 
\def\confName{CVPR}

\title{Automated Natural Language Explanation of \\Deep Visual Neurons with Large Models}

\author{Chenxu Zhao, Wei Qian, Mengdi Huai\\
Department of Computer Science\\
Iowa State University\\
\texttt{\tt\small\{cxzhao, wqi, mdhuai\}@iastate.edu}
\and
Yucheng Shi, Ninghao Liu\thanks{{corresponding author.}} \\
Department of Computer Science\\
University of Georgia\\
\texttt{\tt\small\{yucheng.shi, ninghao.liu\}@uga.edu}
}
\begin{document}
\maketitle
\input{sec/0_abstract}    
\input{sec/1_intro}

\input{sec/2_relatedwork}

\input{sec/3_methodology}
\input{sec/4_experiments}

\input{sec/5_conclusions}
\newpage
{
    \small
    \bibliographystyle{ieeenat_fullname}
    \bibliography{main}
}


\end{document}

%% file: preamble.tex
\usepackage[dvipsnames]{xcolor}


%% file: sec/0_abstract.tex
\begin{abstract}
Deep neural networks have exhibited remarkable performance across a wide range of real-world tasks. However, comprehending the underlying reasons for their effectiveness remains a challenging problem. Interpreting deep neural networks through examining neurons offers distinct advantages when it comes to exploring the inner workings of neural networks. Previous research has indicated that specific neurons within deep vision networks possess semantic meaning and play pivotal roles in model performance. Nonetheless, the current methods for generating neuron semantics heavily rely on human intervention, which hampers their scalability and applicability. To address this limitation, this paper proposes a novel post-hoc framework for generating semantic explanations of neurons with large foundation models, without requiring human intervention or prior knowledge. Our framework is designed to be compatible with various model architectures and datasets, facilitating automated and scalable neuron interpretation. Experiments are conducted with both qualitative and quantitative analysis to verify the effectiveness of our proposed approach. 
\end{abstract}

%% file: sec/1_intro.tex
\begin{figure*}[t]
\vskip -10pt
	\centering
\includegraphics[width=0.99\linewidth]{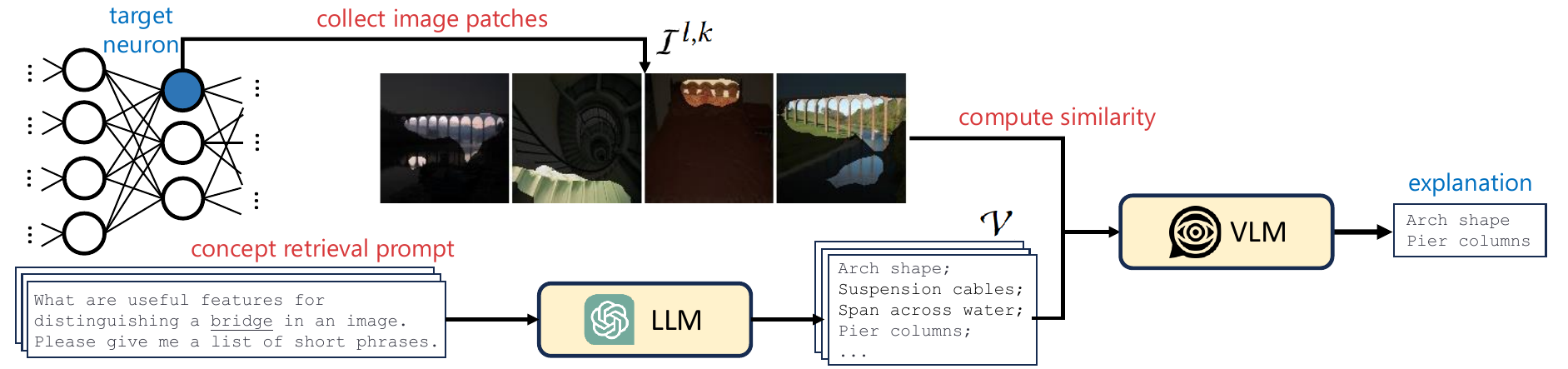}

	\caption{An illustration of the proposed method for explaining a target deep neuron with large models.}

\label{fig:pipeline}
\end{figure*}

\section{Introduction}
\label{sec:intro}

Gaining insights into the behavior of modern machine learning models, particularly deep neural networks (DNNs), remains a significant challenge. 
Interpreting DNNs from the neuron perspective unravels the roles and functions of individual neurons, which has proven to be effective in exploring the inner workings of deep models~\cite{bau2018gan, bau2020understanding, nguyen2016synthesizing, bills2023language, zhou2018revisiting, morcos2018importance}. Some deep neurons are specialized in representing certain semantic concepts, which are crucial in expressing specific classes and making predictions. 

Existing techniques for understanding deep neuron behaviors are still limited. 
Approaches based on visualization~\cite{zhou2014object,bau2017network,zhou2018revisiting,hernandez2022natural,zeiler2014visualizing,olah2017feature,karpathy2015visualizing,zhou2016learning} could identify the relevant features (image regions) of the target neuron, but cannot explain the meaning of the features. This leaves much of the explanation and analysis work to humans, which can lead to a burdensome workload. For example, \citet{zhou2014object}, \citet{bau2017network}, and \citet{hernandez2022natural} all require the engagement of human resources for collecting and labeling new datasets. The involvement of human efforts makes uncovering the semantic meaning of neurons challenging to implement, especially given the fast growing scale of modern models. Hence, there is a strong desire within the research community for an automated method that can provide explanations for neurons with minimal human intervention. 

To address the above challenges, in this paper, we present a novel interpretation approach, which can automatically generate semantic explanations for neurons of DNNs trained on different datasets. It is important to note that our proposed method does not require human prior knowledge, enabling the entire process to be conducted automatically. 
Specifically, we first extract a set of activated image patches that are associated with a given neuron of the target model. In this way, we can identify the specific regions of the input images that contribute to the neuron's activation. 
Then, before generating descriptions for the activated image patches, we construct a vocabulary of semantic concepts that is tailored to the application domain. To obtain a comprehensive vocabulary, we propose prompting large language models to leverage their common-sense knowledge. The vocabulary could constrain the generated description to reduce randomness in explanation results. 
After that, we establish the connection between image patches and the matching concepts. Instead of constructing a dataset to train the annotation model~\cite{hernandez2022natural}, we adopt off-the-shelf vision language models (VLMs) to directly align appropriate concepts with the image patches. By going through the concept vocabulary, we are able to generate a comprehensive and meaningful explanation to understand the target neuron's behavior. 
Our approach is model-agnostic, simple, and flexible, eliminating the need for additional training or manual data collection steps. To validate the efficacy of our proposed method, we conduct a series of experiments. We initially evaluate our method's performance by applying it to the convolutional neural networks~\cite{krizhevsky2017imagenet} and vision transformers~\cite{dosovitskiy2020image} on both the ImageNet~\cite{deng2009imagenet} and Places365~\cite{zhou2017places} datasets. We delve further into exploring the semantic information encapsulated within neurons. Additionally, we carry out a neuron ablation to quantitatively ascertain the significance of these semantic neurons.

%% file: sec/2_relatedwork.tex
\begin{figure*}[htbp]
\vskip -10pt
	\centering
\includegraphics[width=0.99\linewidth]{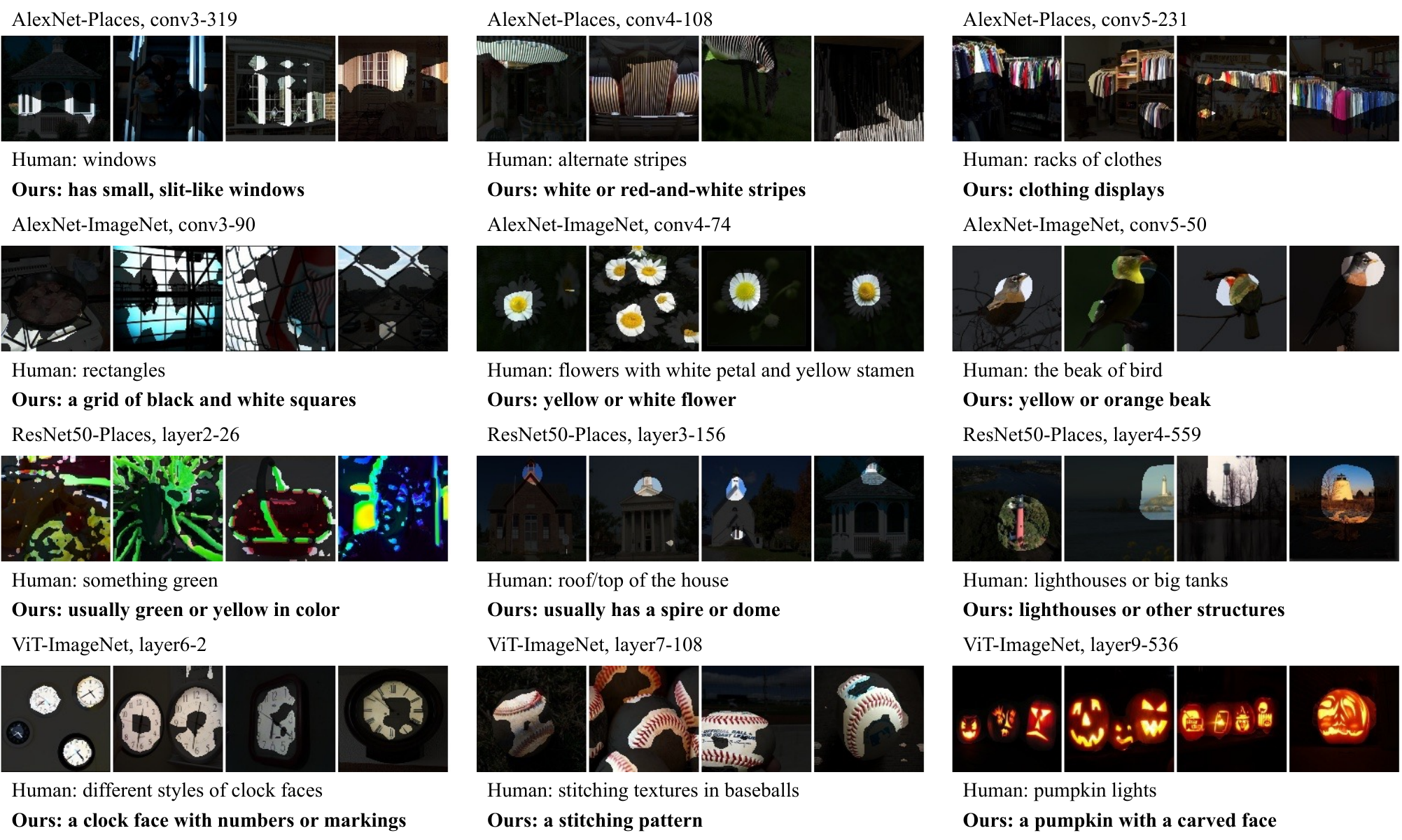}
	\caption{Illustrations of neuron explanations.}
\label{fig:overview}
\end{figure*}

\section{Related Work}
\label{sec:relatedwork}

\noindent \textbf{Explanation Methods.} Existing DNN explanation methods can be categorized based on the level of granularity at which they operate, namely network-level, layer-level, and neuron-level explanation. Network-level explanation~\cite{simonyan2013deep,szegedy2013intriguing,koh2017understanding,murdoch2019definitions,zellers2019recognition} focuses on understanding the overall model behavior given input data, such as gradient-based methods, input perturbations, and influence function. Layer-level explanation~\cite{yosinski2014transferable, selvaraju2017grad, kim2018interpretability, laina2020quantifying, zhang2018interpretable, laina2020quantifying} aims to investigate the activations and learned representations at different DNN layers, illuminating the hierarchical nature of feature extraction and transformation within the model. In contrast, neuron-level explanation methods~\cite{zhou2014object, bau2017network, zhou2018revisiting, hernandez2022natural} focus on understanding the fine-grained functionality of individual neurons. It is also worth noting that our work focuses on post-hoc explanation that can be applied to different types of deep models, instead of developing a certain interpretable architecture~\cite{yang2023language,menon2022visual}.

\noindent \textbf{Neuron Explanation.} The line of neuron-level explanation research has gained substantial attention. A representative study by \citet{zhou2014object} utilize a labor-intensive process, involving human workers from the Amazon Mechanical Turk platform to manually isolate shared features among the top-rated patches for each neuron. The most relevant papers to our current work are  \citet{bau2017network} and \citet{hernandez2022natural}. \cite{bau2017network} assemble a new heterogeneous data set called The Broadly and Densely Labeled Dataset (Broden), which exhibits heterogeneity and pixel-level segmentation for most examples. \cite{hernandez2022natural} hinge on the procurement of a dataset filled with meticulously composed human descriptions, designed specifically for delineating image regions. In stark contrast to these methodologies, which heavily rely on human involvement, our approach distinguishes itself as an autonomous method. It eliminates the need for any form of manual intervention, offering a streamlined and efficient means of extracting neuron-level explanations.

%% file: sec/3_methodology.tex
\section{Methodology}
We consider the image classification model denoted as $f_\theta: \mathcal{X}\rightarrow \mathcal{Y} $ as the target to be explained. Here, $\mathcal{X} \in \mathbb{R}^{D}$ represents the input image space, and $\mathcal{Y} \in [C]=\{1,\cdots,C\}$ represents the class label in the classification task, where $C$ denotes the number of classes. Let $f_\theta^{\smash{l,k}}$ denote the $k$-th neuron in layer $l$. Our goal is to design a novel interpretation method to understand and describe the semantics of these neurons using natural language tokens. We use $\mathcal{W}^{\smash{l,k}} = \{ w_i \}$ to denote the set of tokens that serve as the basis for summarizing the visual concepts captured by neuron $f_\theta^{\smash{l,k}}$, where each token $w_i$ corresponds to a word. 

\textbf{Image Patches Collection.} Here, we present a formal illustration of generating activated patches. Given the target neuron $f_\theta^{\smash{l,k}}$, we construct a set of image patches, denoted as $\mathcal{I}^{\smash{l,k}} = \{\bm{x}^{\smash{l,k}}_p\}$, where each  $\bm{x}^{\smash{l,k}}_p\in \mathcal{X}$ denotes an image patch. These patches are carefully chosen based on their ability to elicit significant activations in the target neuron $f_\theta^{\smash{l,k}}$. In detail, we assume a set of image samples $\mathcal{I} = \{\bm{x}_n\}^{\smash{N}}_{n=1}$ is available, and will construct the patches from these samples. Inspired by~\citet{zhou2014object}, we use a reverse engineering approach to identify the key input regions that cause the activation of neurons. Specifically, we first choose the $K$ images in $\mathcal{I}$ with the highest neuron activations to form the image subset $\mathcal{I}_K$. Then we replicate each image $\bm{x}_n \in \mathcal{I}_K$ for $M$ times, and apply occlusions to the replicated images at different locations using a stride of $3$. As a result, we obtain a set of occluded images denoted as  $\{\bm{x}_n^{\smash{m}}\}_{m=1}^{M}$. These occluded images are then fed into the same network $f_\theta$ to observe changes in the activation of the target neuron $f_\theta^{\smash{l,k}}$. We define discrepancy scores as the neuron activation difference between the occluded images and the original image, i.e., $|f_\theta^{\smash{l,k}}(\bm{x}_n^m)- f_\theta^{\smash{l,k}}(\bm{x}_n)|$, where a higher discrepancy score means the occluded region is more crucial. A neuron receptive field is synthesized as the average of occlusions weighted by the discrepancy scores. This receptive field serves as the basis for generating the activation mask $\bm{m}^{\smash{l,k}}_n$. The mask is then applied to the original input image $\bm{x}_n$, resulting in the creation of a neuron-correspondent image denoted as $\bm{x}^{\smash{l,k}}_p = \bm{x}_n \odot \bm{m}^{\smash{l,k}}_n$. Thus, we can get the set of activated image patches $\mathcal{I}^{\smash{l,k}} = \{ \bm{x}^{\smash{l,k}}_p \}$ for given neuron $f_\theta^{\smash{l,k}}$. 

\textbf{Explanation Vocabulary.} After obtaining the activated patches as representatives of the target neuron $f_\theta^{\smash{l,k}}$, we use natural language to describe the key patterns in those patches, serving as the explanation of $f_\theta^{\smash{l,k}}$. 
To reduce randomness in the explanation, we propose to pre-define a vocabulary, denoted as $\mathcal{V}$, based on the given dataset and scenario.
Each vocabulary item refers to a high-level \textit{concept}, i.e., a word token, a phrase, or a short sentence.
In this work, we let the concepts in $\mathcal{V}$ consist of the feature descriptors obtained by querying an external Large Language Model (LLM) with each of the image labels $y\in \mathcal{Y}$. 
Specifically, we utilize the Generative Pre-trained Transformer (GPT) model for this purpose. The GPT series \cite{radford2018improving,radford2019language,brown2020language,OpenAI2023GPT4TR} serve as an extremely powerful tool, which has demonstrated remarkable proficiency in complex Natural Language Processing (NLP) tasks like machine translation, sequence labeling, reading comprehension, Q\&A \cite{floridi2020gpt, wang2021want, hendy2023good}. Furthermore, researchers have recently recognized and explored the remarkable capabilities of the GPT model, leading to its application in diverse research domains \cite{budzianowski2019hello, hu2020gpt, he2022galaxy, lund2023chatting}. Motivated by this, our methodology involves utilizing the GPT model by prompting each class individually to generate corresponding feature descriptors. By integrating these generated features, we can effectively generate an outstanding concept vocabulary for our purposes. 
For example, we query GPT with a prompt ``What are useful features for distinguishing a \underline{\smash{class name}} in an image? Please give me a list of short phrases.'' to retrieve relevant features for the class. With this approach, a distinct set of concept-specific vocabulary $\mathcal{V}$ can be acquired for the provided dataset.

\textbf{Explanation Generation.} With the activated image patches $\mathcal{I}^{\smash{l,k}}$ and the pre-defined concept vocabulary $\mathcal{V}$, we leverage them to construct explanation as a set of descriptions $\mathcal{W}^{\smash{l,k}}$ for the target neuron $f_\theta^{\smash{l,k}}$. To achieve this, we innovatively incorporate the Vision Language Models (VLMs) such as Contrastive Language–Image Pre-training (CLIP) model~\cite{radford2021learning} into the analysis of neurons. The CLIP model has demonstrated a distinct capability in encompassing a wealth of diverse and comprehensive information and quantifying similarities between arbitrary combinations of data and images through the utilization of an extensive dataset comprising 400 million pairs of image and text data obtained from the internet~\cite{shen2021much, patashnik2021styleclip, tevet2022motionclip}. Therefore, we leverage CLIP to provide an enriched and precise semantic interpretation of the image patches that are most likely to activate a given neuron. 
Specifically, we utilize the CLIP model to compute the similarity scores between the patches in $\mathcal{I}^{\smash{l,k}}$ and each element of $\mathcal{V}$. Based on these similarity scores, the explanation is defined as:
\begin{align}
\mathcal{W}^{\smash{l,k}} &= \{v | v \in \mathcal{V}, \\
&\phantom{=} v \text{ is the top-scoring item according to } s(\mathcal{I}^{\smash{l,k}}, v) \} \nonumber ,
\end{align}
where $s(\mathcal{I}^{\smash{l,k}}, v) = \frac{1}{|\mathcal{I}^{\smash{l,k}}|}\sum_{x_p\in \mathcal{I}^{\smash{l,k}}}\phi(x_p, v)$, and $\phi(x_p, v)$ measures the similarity between image patch $x_p$ and text $v$ according to the VLM $\phi$. Thus, we identify the descriptors from $\mathcal{V}$ that are most likely to align with the image patches. 
Since those patches represent the target neuron $f_\theta^{\smash{l,k}}$, we finally obtain the natural language explanation for our target.

%% file: sec/4_experiments.tex
\section{Experiments}
\begin{figure}[t!]

\centering
\includegraphics[width=0.99\linewidth]{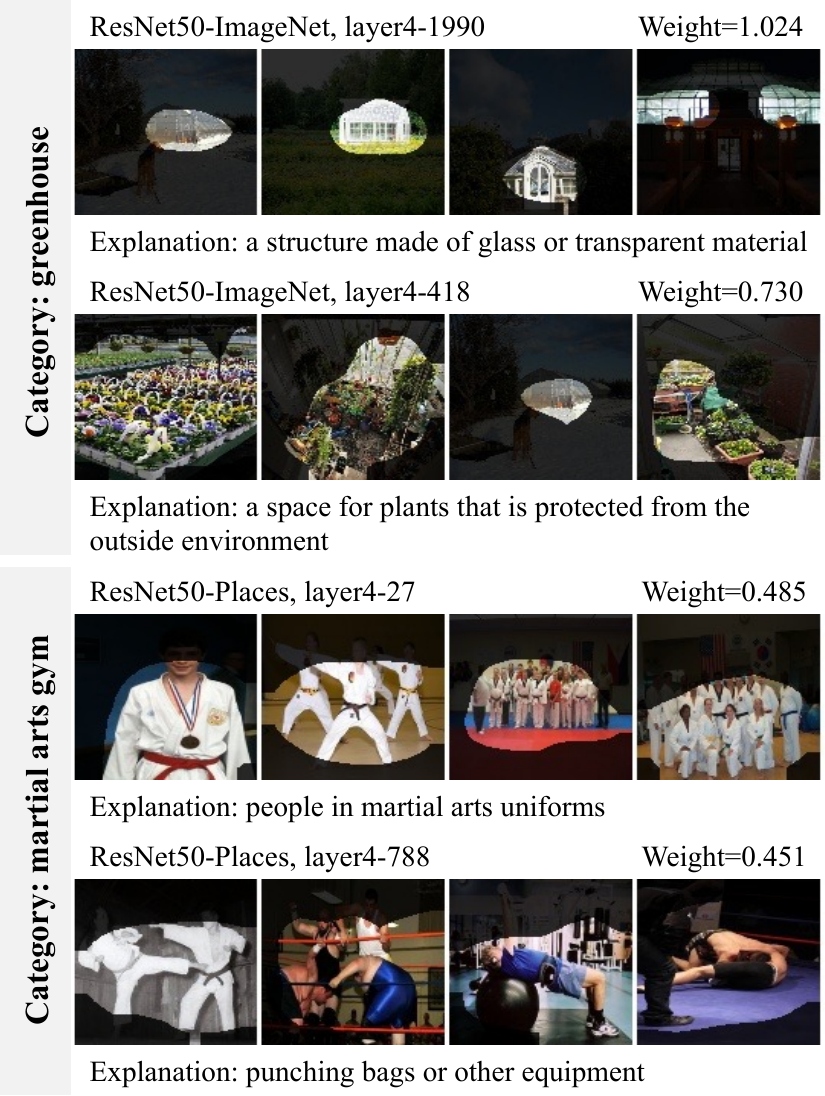}

	\caption{Neuron semantics in categories.}

\label{fig:semantic}
\end{figure}

\subsection{Experimental Settings}

\textbf{Datasets and models.} In experiments, we adopt the following datasets: ImageNet \cite{deng2009imagenet} and Places365 \cite{zhou2017places}. ImageNet is an image classification dataset that contains approximately 1.3 million images from 1000 object categories. Places365 is a scene recognition dataset that contains about 1.8 million images covering 365 scene categories. For the adopted datasets, we analyze and compare the interpretability of neurons within the convolutional layers of ResNet50 \cite{he2016deep} and AlexNet \cite{alex2012deep}, as well as the neurons within the MLP layers of Vision Transformers (ViT) \cite{dosovitskiy2020image}.

\textbf{Baseline.} To compare the interpretability of neurons in deep neural networks, we have five human observers to look at each top-activating image patch and ask them to describe common features or patterns in phrases.

\begin{figure}[t!]

\centering
\includegraphics[width=0.99\linewidth]{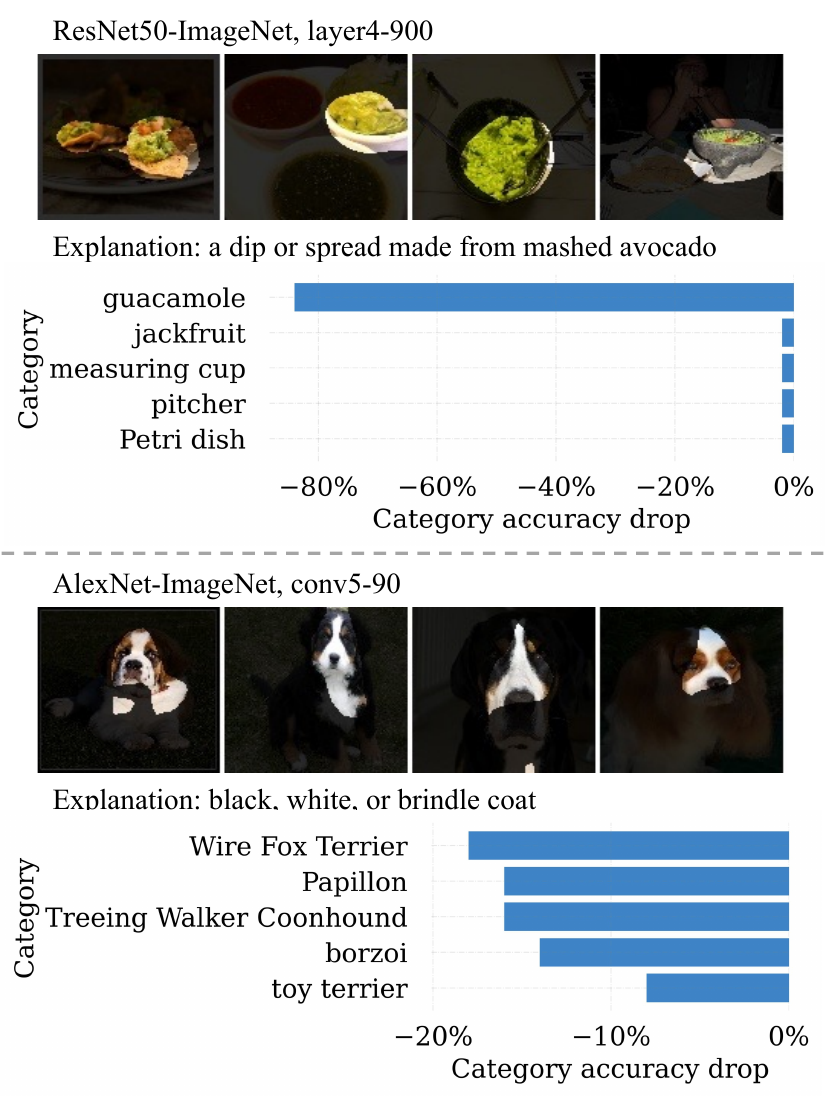}

	\caption{Test accuracy drop after ablating a unit.}

\label{fig:ablated_unit}
\end{figure}

\subsection{Performance of Neuron Explanations}

We begin by evaluating the performance of neuron explanations produced by our proposed method in various layers in deep neural networks. We query the GPT-3 for the feature descriptions from categories in ImageNet and Places365. We utilize the ViT-B/32 model in CLIP to label the neurons' descriptions for the activated image patches. Figure \ref{fig:overview} shows the experimental results on the pre-trained models, namely AlexNet-Places, AlexNet-ImageNet, ResNet50-Places, and ViT-ImageNet. The patch generated by each unit \footnote{We use unit to refer to the neuron with the largest activation in feature maps.} is shown on four maximally activating images. Firstly, we compare the neuron explanations produced by our proposed method with the human baseline. As we can see, our neuron explanations demonstrate a high level of agreement with human annotations. For instance, both descriptions for unit 231 in conv5 of AlexNet-Places depict clothing objects. In some instances, our neuron explanations even provide more detailed feature descriptions than human annotations. For example, for unit 2 in layer6 of ViT-ImageNet, we accurately describe ``a clock face with numbers or markings'' as common features in the given image patches. Secondly, by comparing different layers within the models, we can observe that the units in different convolutional layers are responsible for detecting different levels of patterns or features. Specifically, units in earlier layers usually capture low-level features like colors, edges, or simple textures, such as the ``squares'' in conv3 of AlexNet-ImageNet. As the network progresses deeper, the units tend to capture more  abstract and generic visual features, such as the ``beak'' in conv5 of AlexNet-ImageNet. Overall, our proposed method demonstrates the capability to generate comprehensive descriptions without relying on human annotations, thereby providing efficient and effective explanations for neurons in deep neural networks.

\subsection{Neuron Semantics in Categories}

In this section, we investigate the semantic information embedded in neurons and their significance in categories. We adopt ResNet50-ImageNet and ResNet50-Places models and focus on the last convolutional layer. We generate category-specific units based on softmax weights and determine the units with higher weights for certain categories, indicating their importance in the prediction. Figure~\ref{fig:semantic} provides examples of categories with two units of the highest weights among convolutional units and their neuron explanations. We can observe that those units contribute the most to predicting the category, as they capture the crucial features. Furthermore, the produced neuron explanations better facilitate our understanding of the semantic information embedded in these units. For instance, when examining the output for the category ``greenhouse'', we find that one explanation describes it as ``a structure made of glass or transparent material'', while another explanation characterizes it as ``a space for plants that is protected from the outside environment''. These explanations align closely with the meaning of the category, providing highly valuable interpretability. Therefore, our approach reveals the semantic information captured by different neurons in the layer, contributing to a better understanding of how the models interpret and recognize different categories.

\subsection{Feature Importance Analysis}
Previous sections have qualitatively shown that neuron explanation is closely related to class categories. Now we quantitatively validate this observation by measuring the influence of neurons on model predictions. We conduct experiments to ablate a unit in the last convolutional layer, where semantic information emerges most, and we measure the resulting decrease in category accuracy. To ablate a unit, we set the weights and biases of its feature maps to 0, thereby eliminating its contribution to predictions for any input image. Figure \ref{fig:ablated_unit} presents examples of units with valuable semantic information and illustrates their impacts on category accuracy. We can see that ablating a single unit leads to a significant category accuracy drop for specific categories. For instance, in the first example, we identify a unit that excellently represents ``a dip or spread made from mashed avocado''. However, after ablating this unit, the test accuracy of the ``guacamole'' category suffers a significant drop of $84\%$. Moreover, we compute the max category accuracy drop for all units in layer4 of ResNet-ImageNet and conv5 of AlexNet-ImageNet. The results for the top 256 sorted units are shown in Figure \ref{fig:ablated_unit_max}, providing an overview of each unit's importance to individual categories. These findings highlight the relative importance of neurons that capture attributes and relational features in influencing model behaviors. Particularly, they play a crucial role in recognizing specific subsets of categories within the dataset.

%% file: sec/5_conclusions.tex
\begin{figure}[t!]
\centering
\begin{subfigure}{0.49\linewidth}
\includegraphics[width=1\linewidth]{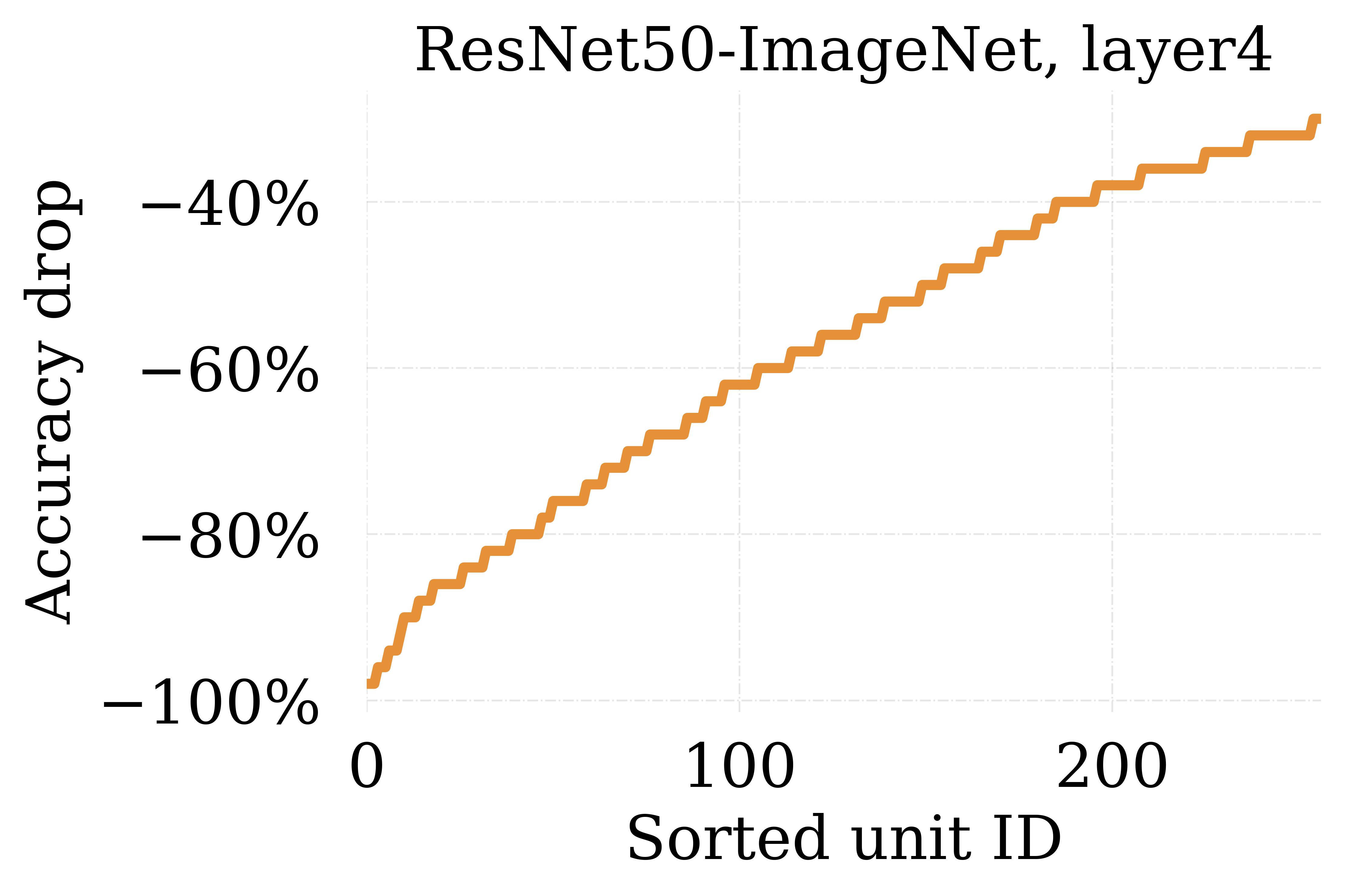}

\label{figs:albated_resnet_max}
\end{subfigure}
\begin{subfigure}{0.49\linewidth}
\includegraphics[width=1\linewidth]{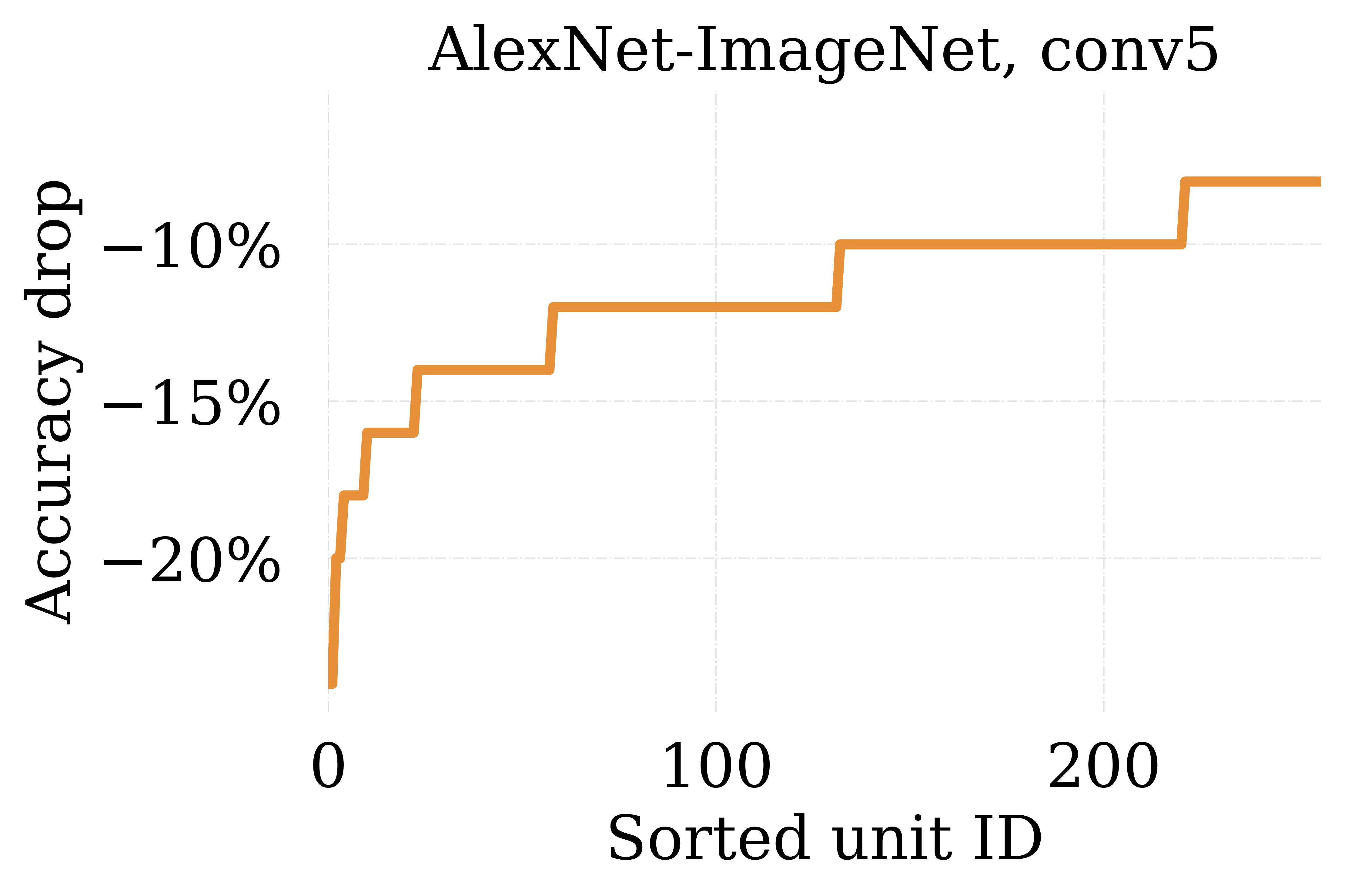}

\label{figs:albated_alexnet_max}
\end{subfigure}

\caption{Max category accuracy drop after ablating a unit. }\label{fig:ablated_unit_max}

\end{figure}

\section{Conclusions}
This paper introduces a novel post-hoc method for generating semantic explanations for neurons within trained deep neural networks. The proposed method eliminates the need for human intervention and can be applied across various deep learning architectures and datasets without limitations. Extensive experiments have been conducted to verify the effectiveness of the proposed method. It is believed that this approach will serve as a valuable tool for the research community, enabling investigations into the mysteries surrounding individual neurons within deep learning models.